\pdfoutput=1
\documentclass[11pt]{article}
\usepackage[]{acl}
\usepackage{acl}
\usepackage{times}
\usepackage{latexsym}
\usepackage[T1]{fontenc}
\usepackage[utf8]{inputenc}
\usepackage{microtype}
\usepackage[none]{hyphenat}
\usepackage{hyperref}
\usepackage{amsmath}
\usepackage{cleveref}
\usepackage{enumitem}
\usepackage{latexsym}
\usepackage{titlesec}
\usepackage{xcolor}
\usepackage{graphicx}
\usepackage{subcaption}
\usepackage{booktabs}
\usepackage{array}
\usepackage{multirow}
\usepackage{tcolorbox}
\title{How Well Do You Know Your Audience? \\
Toward Socially-aware Question Generation}
\author{Ian Stewart \and
Rada Mihalcea \\ 
Computer Science and Engineering \\
University of Michigan \\
\texttt{\{ianbstew,mihalcea\}@umich.edu}
}

\begin{document}

\maketitle

\begin{abstract}
When writing, a person may need to anticipate questions from their audience, but different social groups may ask very different types of questions.
If someone is writing about a problem they want to resolve, what kind of follow-up question will a domain expert ask, and could the writer better address the expert's information needs by rewriting their original post?
In this paper, we explore the task of \emph{socially-aware} question generation.
We collect a data set of questions and posts from social media, including background information about the question-askers' social groups.
We find that different social groups, such as experts and novices, consistently ask different types of questions.
We train several text-generation models that incorporate social information, and we find that a discrete social-representation model outperforms the text-only model when different social groups ask highly different questions from one another.
Our work provides a framework for developing text generation models that can help writers anticipate the information expectations of highly different social groups.
\end{abstract}

\section{Introduction}

% \todo{
% \begin{itemize}
%     \item better motivation (actual example of needing to rewrite for audience + wasting time; education/learning how to do new skill from experts vs. novices)
%     \item more data info: topical diversity
% \end{itemize}
% }

Writers are often expected to be aware of their audience~\cite{park1986} and to minimize the effort required for others to understand them, especially if they cannot receive immediate feedback.
However, NLP tools for writing assistance are not often made aware of the \emph{social} composition of the audience~\cite{ito2019,zhang2020} and the information needs that different people may have.
Preemptive writing feedback may therefore fail to help writers address the expectations of different people in their audience.
% For instance, someone who is writing about their finances may omit an important detail that financial novices may not find relevant (e.g., their location), but which experts may need to know in order to respond with an informed opinion.
This is especially important when the writer requests feedback from a specific group of people:
in one post on a forum related to personal finance, a writer asks for help from financial ``gurus'' for advice about accepting a job offer.
% shown by a post on Reddit about buying computer parts that requests feedback from ``more experienced [PC] builders who can tell me the smartest thing to do.''
% \todo{what is best example case to show value of audience-specific questions?}
% other examples: ``im really just looking for advice from those similar to me maybe, what should i be doing?'' ``How much do you have saved in your retirement plans?'' (EXPERT)
% data_processing/look_for_valid_questions_in_comments.ipynb#Look-for-posts-that-ask-for-social-groups
% This is particularly important for writers when they request additional information to solve a problem in their life, such as a financial matter that is beyond their expertise.

A system that can preempt the hypothetical audience's information needs would enable the writer to revise their original post and avoid possible information gaps~\cite{liu2012}.
Some online forums have already implemented crude solutions for this problem with automated reminders for writers to include basic information (e.g., location) in their post.
Providing writers with preemptive questions can help especially in domains where different social groups have diverse information expectations.
% , such as people sharing news online about an emerging topic for which different social groups expect different kinds of information.
% For example, many people turn to online forums to seek advice~\cite{govindarajan2020}, and generating socially-specific questions would be especially useful for advice-seekers who want to connect with specific audience groups.
In the earlier example about personal finance, the advice-seeker could adapt their original post with answers to hypothetical ``expert-level'' questions (e.g. ``Have you saved enough money for retirement?''), adding extra information that would enable experts to provide advice more quickly.

% \todo{motivation: we want models that expose divergent responses to the same content, to help writers prepare for audience differentiation; connect to generation evaluation}

We cannot predict everyone's information needs, but some social groups with similar backgrounds (e.g., domain experts) will likely have consistent patterns in information expectations~\cite{garimella2019,welch2020}.
In this work, we evaluate several \emph{socially-aware} question generation models with the goal of providing customized clarification questions to writers.
% This study does not seek to maximize overall model accuracy but rather to identify the types of questions and social representations that best suit the task of socially-aware question generation.
% We argue that question generation models should produce \textit{diverse} output that can cover a broad range of possible social needs~\cite{zhang2021}.
% This kind of model could prove useful for people learning how to write in new contexts, in which their audience likely contains people with different levels of domain expertise and interest.

% \input{example_generated_questions}

% We provide examples of questions generated by one of the audience-aware models in \autoref{tab:example_generated_questions}.
% The post seeks advice about installing a graphics card on a computer, and the reader-aware model accurately generates a ``novice'' question about connecting to the GPU (a seemingly obvious detail to request), while also being able to predict an ``expert'' question about installing driver software.

Our work contributes answers to the following questions:

\begin{itemize}[noitemsep, wide, labelwidth=!, labelindent=0pt]
    \item \textbf{How different are social groups based on the questions that they ask?} We collect a dataset of 200,000 Reddit posts seeking advice about a variety of everyday topics such as technology, legal issues, and finance,  containing 700,000 questions.\footnote{We will release the IDs for the post and author data, as well as the data processing code, to aid replication.}
    We define several social groups that are relevant to possible information expectations such as expertise (\autoref{sec:social_background}). 
    We demonstrate that different social groups, e.g. experts vs. novices, ask consistently different questions (\autoref{sec:compare_social_groups}, \autoref{sec:qualitative_analysis_model_output}).
    \item \textbf{How well can generation models predict socially-specific questions?} We extend an existing generation model to incorporate social information about the question-askers (\autoref{sec:model_design}).
    % We introduce a framework to incorporate discrete social representations (defined in \autoref{sec:reader_background}) into a token-based model (\autoref{sec:reader_tokens}) and an attention-specific model (\autoref{sec:reader_attention}).
% We also detail reader-specific embeddings to use as a continuous representation of $r$ (\autoref{sec:reader_embeddings}).
    %The reader-attention model achieves equal performance relative to the text-only baseline (\autoref{sec:question_generation}), while the continuous-representation reader models tend to do worse.
    % We develop several evaluation strategies to highlight desired qualities of socially-aware question generation.
    In automated evaluation, a token-based socially-aware model outperforms the baseline for questions that are ``divisive'' and questions that are specific to a social group, particularly with respect to location as a social group (\autoref{sec:divisive_posts}, \autoref{sec:generation_social_specific_questions}).
    % The socially-aware generation models are therefore best-suited to posts with information expectations that are specific to certain groups (\autoref{sec:question_generation}).
    \item \textbf{Are socially-aware questions useful for writers?} In human evaluations, we found that the socially-aware model is preferred over the text-only model for questions related to the question-asker's location and within the general advice-seeking domain (\autoref{sec:generation_annotation}). This reinforces the utility of socially-aware models in scenarios where the social information is well-defined and where the topics are related to everyday concerns.
    % (; \autoref{sec:generation_annotation}).
    % Furthermore, in human evaluation one of the reader-aware models performs at-par or better than the text-only model.
    % \item We conduct a human evaluation to determine both the relative difference of reader-specific questions and the impact of the questions on the writers' output. We find that the reader-aware model generates questions that are slightly more answerable and understandable than the text-only model.
\end{itemize}

Importantly, the research presented in this paper shows that there are significant differences across groups with respect to questions they ask, and that we can develop models that are more attuned to these differences.
Note that the goal of our work is not to improve the overall accuracy of a question generation system, but rather to develop methods that are sensitive to the needs of specific groups, thus paving the way toward technology that is available and useful for all.

\section{Related Work}

\paragraph{Question generation}

Question generation (QG) is unique among text generation tasks because it tries to address what a person \emph{does not know}, rather than what they already know and want to write.
QG systems are expected to create fluent and relevant questions based on prior text, in order to provide QA systems with augmented data~\cite{dong2019} and students with question prompts to help their learning~\cite{becker2012,liu2012}.
% Typical approaches have extended existing sequential generation language models, such as the LSTM and the transformer, to accommodate both the textual context (e.g. a document) and the likely answer to the question, in order to maximize the probability of generating an informative question~\cite{du2017,indurthi2017}.
In addition to typical supervised learning approaches~\cite{du2017}, reinforcement learning has proven useful, where questions are assigned a higher ``reward'' if they are more likely to have interesting answers~\cite{qi2020} and more relevant to the context~\cite{rao2019}.
Furthermore, work such as \citeauthor{gao2019} (\citeyear{gao2019}) has proposed \emph{controllable} generation techniques to encourage less generic questions, e.g. with higher difficulty.
Such controllable-generation systems often leverage human-generated questions from a variety of domains, including Wikipedia~\cite{du2018}, Stack Overflow~\cite{kumar2020}, and Twitter~\cite{xiong2019}.

To our knowledge, prior work in question generation did not leverage the prior expectations of the question-askers.
While sometimes providing controls for difficulty, no datasets currently include information about the inferred \emph{background} of the question-askers.
It seems natural that a person's prior knowledge would shape the information that they seek in response to a particular situation, yet analysis of the impact of social information on question generation remains absent.
% Furthermore, QG data often focuses on factual domains such as Wikipedia that are unlikely to reveal differences in subjective experience among question-askers.
% Finally, most data sets only focus on questions whose answers are contained in a given document, while we are interested in information that the document lacks~\cite{majumder2021}.
This study tests the role of social information in question generation using a dataset of posts from online forums, which feature complicated scenarios that can result in different information expectations between social groups.
% [TODO: generation for better QA]
% [TODO: generation for (potentially) helping writers]

% [TODO: NLP writing tools, suggestions]
% \vspace{-10pt}

\paragraph{Language model personalization}

Personalized language modeling often seeks to improve the performance of common language tasks, such as generation, using prior knowledge about the text's author~\cite{paik2001}.
Personalization can improve task performance and make language processing more human-aware~\cite{hovy2018}, which ensures that a more diverse population is included in language models~\cite{hovy2016}.
% While initially focused on recommendation and generation, human-aware language processing has proven useful to a variety of prediction tasks such as syntactic parsing~\cite{garimella2019} and geolocation~\cite{bamman2014}.
To represent the text writer, personalized systems often integrate a writer's identity~\cite{welch2020} or a writer's social network information~\cite{del2019} into existing language models.
A more generalizable approach converts the text-writer to a latent social representation such as an embedding~\cite{pan2019}, to be combined with the language representation in a neural network model where the social and text representations are learned jointly~\cite{miura2017}.
We draw inspiration from the \emph{contextualized} view of personalization from \citeauthor{flek2020} (\citeyear{flek2020}), and we represent the question-askers based on their prior behavior with respect to the specific \emph{context} of a given post.

% [TODO: generating news comments, ranking comments (DNM etc.)]

\section{Data}
\label{sec:data}

In this study, we consider the task of generating clarification questions on information-sharing posts in online forums.
% Prior work in question generation has relied on publicly available data such as Wikipedia~\cite{du2018} and StackOverflow~\cite{kumar2020}, which are large-scale and highly diverse but lack prior information about the question-askers.
% In light of the relative lack of \emph{personalized} question data, we collect a new data set from online forum posts that includes non-trivial inference about author identity.
% Prior work in social computing has illustrated the rich diversity of advice-seeking behavior on social media such as Reddit~\cite{fu2019,govindarajan2020,lahnala2021}, which leads us to focus on several popular advice-seeking forums on Reddit.
We choose to study subreddits that have a high proportion of text-only posts, diverse topics, and where community members often ask information-seeking questions: \texttt{Advice} (lifestyle improvement), \texttt{AmItheAsshole} (social norms in complicated situations), \texttt{LegalAdvice} (law disputes), \texttt{PCMasterRace} (computer technology), and \texttt{PersonalFinance} (money and investment).
% \footnote{\texttt{AmITheAsshole} hosts discussions about social norms in everyday situations; \texttt{Advice} hosts discussions about lifestyle choices including work, health, and family.}
% We show basic statistics for these subreddits' submissions and comments in... [TODO: basic statistics (total submissions/comments, post length, rate of question-asking)].\footnote{The data were collected from PushShift~\cite{baumgartner2020} in November 2020.}
We collect all submissions ($\sim$ 8 million) to the above subreddits from January 2018 through December 2019, using a public archive~\cite{baumgartner2020}.
We filter the post data to only include submissions written in English with at least 25 words, which we chose as a cutoff for posts that lack the context necessary for people to ask informed questions.
To identify potential clarification questions, we collect all the comments of the submissions ($\sim$ 6 million) that are not written by bots, based on a list of known bot accounts like \texttt{AutoModerator}.

% \subsection{Identifying valid questions}

% \todo{make this shorter}

% \todo{add to supplementary material, it slows down the paper}
We conduct extensive filtering to include questions that are relevant and that seek extra information from the original post.
The details are available in \autoref{sec:data_question_filtering}.
We summarize the overall data in \autoref{tab:data_summary_statistics}, and we show the distribution of the posts and questions among subreddits in \autoref{tab:subreddit_data_summary_statistics}.
Example posts and associated clarification questions are shown in \autoref{tab:social_representation_data}.
% (see ``Example question'' and ``Example post title'').

\begin{table}[t!]
\centering
\small
\begin{tabular}{>{\raggedright\arraybackslash}p{5.5cm} r}
~ & ~ \\ \toprule
Total posts & 270694 \\
Total questions & 730620 \\
Post length & 304 $\pm$ 221 \\
Question length & 13.9 $\pm$ 8.08 \\
Questions with question-asker data & 77.7\% \\ 
Questions with discrete question-asker data & 75.2\% \\ 
Questions with question-asker embeddings & 43.5\% \\ \bottomrule
\end{tabular}
\caption{Summary statistics about posts, questions, and question-asker data.}
\label{tab:data_summary_statistics}
% scripts/data_processing/compute_data_summary_statistics.ipynb
\end{table}

\begin{table}[t!]
\small
\centering
\begin{tabular}{l r r}
Subreddit & Posts & Questions \\ \toprule
\texttt{Advice} & 48858 & 87592 \\
\texttt{AmItheAsshole} & 61857 & 331345 \\
\texttt{LegalAdvice} & 53577 & 92737 \\
\texttt{PCMasterRace} & 31657 & 47613 \\
\texttt{PersonalFinance} & 74745 & 171333 \\ \bottomrule
\end{tabular}
\caption{Summary statistics about subreddits.}
\label{tab:subreddit_data_summary_statistics}
\end{table}

\section{Defining social groups}
\label{sec:social_groups}

% The typical goal of question generation is to accurately predict the text of question $q$ considering the context of a post $p$.
In this work, we assess the relevance of the question-asker's background in the task of question generation, by defining social groups and assessing their differences in question-asking.
% how can we best capture the prior knowledge and the likely interests of reader $r$ with respect to post $p$, to better predict the reader's question $q$?
% Specifically, we develop models that take as input both the text of a post $p$ and some representation of the post's reader $r$, in order to generate the reader's question $q$.
% The goal is to find model parameters $\theta$ that maximize the conditional likelihood of observing $q$ given $p$ and $r$:
% $$\hat{\theta} = \argmax_{\theta} P(q | p, r, \theta)$$
% This task is different from other settings that condition on the question \emph{answer}~\cite{du2017}.

% stinky old crowded table
% \begin{table*}[h!]
% \centering
% \begin{tabular}{l p{7cm} p{4cm}}
% Reader representation & Value & Example questions \\ \hline
% Prior topic interest & \% of prior comments in subreddit where reader is currently commenting. & \\ \hline
% Demographic & Self-reported location (country-level). & \\ \hline
% Response time & Time between original post and reader's comment. & \\ \hline 
% Prior topic distribution & Embedding of subreddits where reader commented previously. & TODO: example questions from different categories of reader \\ 
% ~ & Embedding of text in reader's previous comments. & \\ \hline
% \end{tabular}
% \caption{Variables to represent readers in question generation system.}
% \label{tab:reader_representation_data}
% \end{table*}

% \usepackage{multirow}
\begin{table*}[t!]
\scriptsize
\centering
\begin{tabular}{l p{3cm} p{2cm} >{\raggedright\arraybackslash}p{4cm} p{3cm}}
Group category & Description & Social group & Example question & Example post title \\ \toprule
\texttt{EXPERTISE} & \multirow{2}{3cm}{Prior rate of commenting in the target subreddit, or a topically-related subreddit.} & \texttt{Expert} \newline ($\geq$ 75th percentile) & How much would you need to make on day 1 to meet your current financial obligations? & \multirow{2}{3cm}{(\texttt{PersonalFinance}) Changing careers at 39} \\ \addlinespace[0.15cm]
 &  & \texttt{Novice} \newline ($<$ 75th percentile) & Where do you live? &  \\ \midrule
\texttt{TIME} & \multirow{2}{3cm}{Mean amount of time elapsed between original post and question-asker's comment, among all prior comments.} & \texttt{Fast} \newline ($<$ 50th percentile) & Does your wife have a relationship with him? & \multirow{2}{3cm}{(\texttt{LegalAdvice}) Having a child and partner's father is sex offender} \\ \addlinespace[0.15cm]
 &  & \texttt{Slow} \newline ($\geq$ 50th percentile) & If he is a sex offender, shouldn't he be kept away from children? &  \\ \midrule
\texttt{LOCATION} & \multirow{2}{3cm}{Inferred location of question-asker.} & \texttt{US} & Have you looked at the RX 580? & \multirow{2}{3cm}{(\texttt{PCMasterRace}) Should I buy GTX 1050Ti?} \\ %\addlinespace[0.7cm]
 &  & \texttt{non-US} & The 1050 is 160\$ in India? & \\ \bottomrule
\end{tabular}
\caption{Group categories for question-askers, with example questions and posts.}
\label{tab:social_representation_data}
% scripts/data_processing/compare_reader_group_questions.ipynb#Get-example-questions-from-different-groups
\end{table*}

\subsection{Defining social groups}
\label{sec:social_background}

% In this study, we assume that the post writer and question-asker (``reader'') have no prior experience with one another and are attempting to exchange information rationally (as opposed to e.g. joking, trolling).
% This kind of situation reflects the Rational Speech Act framework~\cite{grice1975} in which speakers cooperate to reach a shared understanding of the world.
% We assume that the prior background of a question-asker plays a role in their information-seeking goals.
We collect a limited history for the question-askers ($N=1000$ comments) to quantify relevant aspects of their background that may explain their information-seeking behavior.
% When considering their readers, a writer may not have perfect knowledge of the readers' backgrounds, their goals, or their possible stance toward the topic of discussion.
% However, a writer may be helped by making assumptions about the readers' status \emph{relative} to the topic of discussion, e.g. if the reader is highly experienced in the subject matter.
% While there are many possible social attributes that can affect question asking, 
% Out of many possible social attributes, this study addresses attributes that are readily extracted from text and cover a wide variety of readers.
We consider the following social groups who are likely to have different information expectations:

\begin{enumerate}[noitemsep, wide, labelwidth=!, labelindent=0pt]
    \setlength\itemsep{0em}
    % \item Is the reader highly interested in the topic of inquiry, or are they interested in a variety of topics? A reader with consistent interest in the topic of inquiry may ask a detailed question about a particular aspect of the writer's request, while a reader with more diverse interests might ask a more generic question.
    \item \texttt{EXPERTISE}: 
    % Does the reader have more or less experience in the post topic? 
    A question-asker with less experience may ask about surface-level aspects of the post, while someone with more expertise might ask about a more fundamental aspect of the post.
    We quantify ``expertise'' using the proportion of prior comments that the question-asker made in the subreddit $s$ (or a topically related subreddit; see \autoref{sec:model_topical_subreddits}) in which the original post was made.
    For example, if a question-asker has frequently written comments in \texttt{WallStreetBets} before asking a question in \texttt{PersonalFinance}, they are likely more familiar with financial terms than the average person.
    We define an \texttt{Expert} question-asker as anyone at or above the $75^{th}$ percentile of rate of commenting in a relevant subreddit, and a \texttt{Novice} question-asker as anyone below the percentile,
    % \footnote{For both \texttt{EXPERT} and \texttt{TIME}, we tested other cutoff values and found similar accuracy for the group-classification task.} 
    where we chose the threshold to fit the skewed data distribution.
    Other threshold values produced similar results in social group classification.
    \item \texttt{TIME}: 
    % Does the reader tend to react quickly or slowly to posts? 
    A question-asker who replies soon after the original post was written may ask about missing information that is easily corrected (e.g. clarifying terminology), while a question-asker who replies more slowly may ask about more complicated aspects of the writer's request (e.g. the writer's intent).
    We quantify this with the mean speed of responses of the question-asker's prior comments \emph{relative} to the parent post.
    We define a \texttt{Slow} question-asker as anyone at or above the $50^{th}$ percentile of mean response time, and a \texttt{Fast} question-asker as anyone below the threshold.
    % We quantify this quality based on the relative time elapsed between the original post and the reader's question.
    \item \texttt{LOCATION}: 
    % Does the reader identify with a particulary geographic region?
    A question-asker who is based in the US may ask questions that reflect US-centric assumptions, while a non-US question-asker may ask about aspects of the post that are unfamiliar to them.
    % \footnote{Using ``non-US'' as a single category combines people from many different countries, but we use this category to avoid data sparsity and to group clear patterns among people from other countries (e.g. non-US vocabulary).}
    % shares a similar location to the post author may ask questions that ``skip over'' their shared experiences, while a reader who is from a different location may ask questions that elicit more background information.
    We quantify location with the question-asker's self-identification from prior comments, using Stanza's English NER tool~\cite{qi2020stanza} to identify LOCATION entities and OpenStreetMap to geo-locate the most likely locations.
    % \footnote{A statement such as ``I live in the US'' has ``US'' tagged by an high-precision NER system~\cite{qi2020stanza} and geolocated to a known geographic entity in OpenStreetMap with high confidence.} 
    % and from their posting history in location-specific subreddits.
    For those without self-identification, we identify all location-specific subreddits $\mathcal{S}_{L}$ in a question-asker's previous posts based on whether the subreddit name can be geolocated with high confidence (e.g., \texttt{r/NYC} maps to New York City). 
    A question-asker $a$'s location is identified with the location-specific subreddit where $a$ writes at least 5 comments and where they write the most comments out of all location-specific subreddits $\mathcal{S}_{L}$.
    % A reader with a less similar background may ask questions that relate to aspects of the writer's background, while a reader with a more similar background may not need to ask such questions since they have a shared understanding of what that background entails.
\end{enumerate}

% \vspace{-8pt} % fix spacing

We summarize these definitions of different social groups in \autoref{tab:social_representation_data}.
% We emphasize that these groups are not specific to Reddit and can be reproduced in other domains.
The example questions in demonstrate that question-askers who occupy different groups tend to ask questions about different aspects of the original post: e.g. the \texttt{Fast} question-asker addresses a basic fact about the situation, while the \texttt{Slow} question-asker addresses a more complicated/hypothetical point.
% Note that many question-askers may belong to multiple groups: an \texttt{Expert} asker could also belong to \texttt{non-US}.
% For simplicity, in all our models we split question-askers who belong to multiple group categories ($\sim 37\%$) into different data points.
% num(authors w/ >= 1 category) / num(authors) from combined_data_clean_question_data.gz
% If question-asker $a$ belongs to multiple social groups (37\% of the question-askers), we provide the context post $p$ and the specified question $q$ to the model two times for different groups $g_{1}$ and $g_{2}$.

\subsection{Validating group differences}
\label{sec:compare_social_groups}

\begin{table}[t!]
\small
\begin{tabular}{l >{\raggedright\arraybackslash}p{4cm}}
Group category & Top-3 LIWC categories (absolute frequency difference) \\ \toprule
\texttt{LOCATION} &  \\
\texttt{US} \textgreater \texttt{non-US} & MONEY (0.512\%), WORK (0.361\%), RELATIV (0.337\%) \\ \addlinespace[0.1cm]
\texttt{non-US} \textgreater \texttt{US} & FOCUSPRESENT (0.356\%), FUNCTION (0.327\%), AUXVERB (0.305\%) \\ \hline
\texttt{EXPERTISE} &  \\
\texttt{Expert} \textgreater \texttt{Novice} & MONEY (0.207\%), YOU (0.135\%), FOCUSPRESENT (0.106\%) \\ \addlinespace[0.1cm]
\texttt{Novice} \textgreater \texttt{Expert} & DRIVES (0.097\%), AFFILIATION (0.056\%), REWARD (0.055\%) \\ \hline
\texttt{TIME} &  \\
\texttt{Fast} \textgreater \texttt{Slow} & YOU (0.312\%), PPRON (0.225\%), PRONOUN (0.160\%) \\ \addlinespace[0.1cm]
\texttt{Slow} \textgreater \texttt{Fast} & DRIVES (0.105\%), AFFECT (0.082\%), IPRON (0.066\%) \\ \bottomrule
\end{tabular}
\caption{LIWC category word usage differences across social groups (\% indicates absolute difference in normalized frequency). All differences are significant with $p<0.05$ via Mann-Whitney U test.}
\label{tab:social_group_word_diffs}
% \vspace{-12pt} % how to reduce space below table??
% scripts/data_processing/compare_reader_group_questions.ipynb#Identify-word/phrase-differences-across-groups
\end{table}

As a first step, we test for consistent differences in the types of questions asked by different social groups.
% For each social category, we sample an equal number of members from each group and from each subreddit to avoid biasing toward a particular group or subreddit.
% For every post $P$ that receives multiple questions, we identify question $q_{i}$ from a reader of group $i$ and $q_{j}$ from a reader of group $j$ (e.g. US reader and non-US reader).
% For each social group $g$ in category $\mathcal{G}$ (e.g., groups \texttt{US} and \texttt{non-US} from \texttt{LOCATION}), we sample an equal number 
% $N_{\mathcal{G},s}$ questions from each subreddit $s$, where $N_{\mathcal{G},s}=min\{N_{g,s} \: \forall \: g \in \mathcal{G} \}$.
We test for topical differences between the groups by comparing the relative rate of LIWC word usage in their questions, a common strategy to identify salient differences between social groups~\cite{pennebaker2001}.
The results in \autoref{tab:social_group_word_diffs} show consistent differences in word usage in the questions.
\texttt{Expert} question-askers ask about money more often than \texttt{Novices}, which could indicate an assumption from prior experience that post authors' core problems stem from their financial decisions (even outside of the finance-related subreddits).
Similarly, \texttt{US} question-askers have more questions about money and work than \texttt{non-US} readers, who often frame questions to address present-tense issues and write with more auxiliary verbs.
Fast-response question-askers ask more often about the post author (YOU), which may indicate a stronger interest toward the post author's background, as opposed to slow-response question-askers who address the poster's high-level intentions (DRIVE) and emotional behavior (AFFECT).
While it is possible that some of these differences are spurious, it is unlikely that they all relate to stylistic patterns such as regional differences (\texttt{LOCATION}), considering the prevalence of relevant LIWC categories (e.g. MONEY relates to financial questions, which are relevant to the data).

We verify these differences with a classification task, which we detail in \autoref{sec:social_group_classification}.

% (e.g., high ``YOU'' use, see \autoref{tab:reader_group_word_diffs}).
% We find an unusually high performance for \texttt{LOCATION}, which partly points to stylistic discrepancies between \texttt{US} and \texttt{non-US} readers (e.g. \texttt{non-US} readers use non-American spellings such as ``behaviour'' and ``cheque'').
% scripts/data_processing/compare_reader_group_questions.ipynb#Debugging-data-samples:-overfitting

\section{Question generation}

\subsection{Model design}
\label{sec:model_design}

We build the generation models on top of the BART model~\cite{lewis2020}, a transformer model known to be resistant to data noise.
We use the same pre-trained model (\texttt{bart-base}; $|V| \approx 50,000$) and the same training settings for all models.\footnote{Learning rate 0.0001, weight decay 0.01, Adam optimizer, 10 training epochs, batch size 2, max source length 1024 tokens, max target length 64 tokens, cross-entropy loss.}
% We only tested one type of generation model under the assumption that different models would process the addition of social information in a similar way.
The main point of the model modifications is not to achieve universally high accuracy, but to assess the value of different social data representations in question generation.
% We use cross-entropy loss during training to maximize the conditional probability of generating a question given the provided post.

\subsubsection{Social tokens}
\label{sec:social_tokens}

% In the ``discrete'' approaches, we represent a reader using a specific aspect of their background behavior.
For the ``social-token'' model (a discrete representation), we add a special token $\{GROUP_{g}\}$ to the text input of the baseline model to indicate whether the asker belongs to social group $g$ (cf. prior work in controllable generation; \citeauthor{keskar2019} \citeyear{keskar2019}).
% , which use domain-specific tokens when ``prompting'' the model.
The embeddings for these social tokens are learned during training in the same way as the other text tokens.
All question-askers who could not be assigned to a group are represented with \texttt{UNK} tokens.

% \begin{enumerate}[noitemsep]
%     \item Expert vs. non-expert, based on \% of prior posts in the subreddit $s$ in which the post was made.
%     \item Fast vs. slow response, based on the amount of time elapsed between the post and the question.
%     \item U.S. vs. non-U.S. reader, based on the reader's self-identification from prior posts (e.g. ``I live in X'' where ``X'' is identified by an NER system and geolocated to a real geographic entity via OpenStreetMap).
% \end{enumerate}

\subsubsection{Social attention}
\label{sec:social_attention}

% Some work in personalization has suggested training a separate model for different author groups~\cite{welch2020}, which can require a huge number of parameters.
% An alternative way to represent the ``reader'' within the model is to teach the model to pay attention to different parts of the input according to the reader's likely interests.
For discrete modeling, we also consider customizing a separate part of the model for different social groups.
% The transformer model relies on multiple layers of attention modules~\cite{vaswani2017} to determine the most important part of the input text for the generation task.
Specifically, we change one of the attention layers of the typical transformer model~\cite{vaswani2017} to represent differences in how different question-askers may perceive a post.

We replace attention module $\ell$ in the encoder with a different module for each social group $g$.
% , and we dynamically switch to the module associated with group $g$ when predicting a question generated by reader group $g$.
For regularization, we train a separate \emph{generic} attention module at the same time as the social-group attention, concatenate the social attention with the generic attention, and pass the concatenated attention through a linear layer to produce the final attention distribution.
% We compute the final attention for a post $p$ to which reader $r$ responds as $\hat{W} = \frac{1}{2}(W + \lambda{}W_{r}$). 
We choose the layer index $\ell=1$ from $\{1,3,5\}$
% and $\lambda=0.9$ from $\{0.1, 0.5, 0.9\}$ 
through performance on validation data.
% We also experimented with other modifications, such as computing a weighted average of the attention distributions, and found that concatenation had the best performance.}
For a question written by an asker who belongs to group $g$ ($gen$ indicates generic attention, $f$ indicates a feed-forward linear layer), the attention is computed as follows:
% \vspace{-2pt}

{
\scriptsize
$$
\text{Multihead}_{\ell}(x) =  f([\text{Multihead}_{g}(x); \text{Multihead}_{gen}(x)])
$$
}
\vspace{-30pt}
% \begin{equation}
%     \scriptsize
%     \begin{aligned}
%     % \text{Multihead}_{\ell} = f( \\ \text{Concat}(\text{head}_{1,gen}, ... \text{head}_{h,gen})W^{O}_{gen}, \\
%     % \text{Concat}(\text{head}_{1,g}, ... \text{head}_{h,g})W^{O}_{g})
%     \text{Multihead}_{\ell}(x) =  f(\text{Concat}(\text{Multihead}_{g}(x), \text{Multihead}_{gen}(x)))
%     \end{aligned}
% \end{equation}

% To avoid overfitting to specific reader groups, we only use reader-specific attention in a single layer.

\subsubsection{Social embeddings}
\label{sec:social_embeddings}

% The previous models represent readers as \emph{discrete} entities, e.g., as an ``expert'' or ``novice.''
For a \emph{continuous} approach to personalization~\cite{wu2021}, we represent question-askers using latent embeddings $e^{(a)}$ based on their prior \emph{subreddit} and \emph{text} behavior.

For \emph{subreddit} behavior, 
% we compute an embedding based on the subreddits in which the question-asker commented before writing their question.
we compute the cross-posting matrix $\mathcal{P}$ for all subreddits and all question-askers in our data, where $P_{i,j}$ is equal to the NPMI of question-asker $j$ writing a comment in subreddit $i$.
% ($-\text{log}(\frac{p(i,j)}{p(j)p(i)})$).
% \footnote{Each cell $\mathcal{P}_{i,j}$ is set equal to the PMI of author $j$ commenting in subreddit $i$.}.
We compress the matrix using SVD ($d=100$), 
% \footnote{We chose $d=100$ as the dimensions to balance data heterogeneity with generalization.}, 
and the subreddit embedding $e_{s}^{(a)}$ for question-asker $a$ is set to the average of the embeddings across all subreddits in which $a$ previously posted.
% $\text{history}$, i.e. 
% $e_{s}^{(a)} = \frac{1}{N} \sum_{s \in \text{history}(a)} \hat{\mathcal{P}}_{s}$.
For \emph{text}, we compute an embedding based on the question-asker's previous comments.
We train a \texttt{Doc2Vec} model $\mathcal{D}$~\cite{le2014} on all prior comments and represent each comment as a single document embedding ($d=100$, default skip-gram parameters).
% We choose \texttt{Doc2Vec} because it has been shown to be useful in capturing general topical trends among variable-length social media posts~\cite{curiskis2020}, which is useful in capturing the background and interests of the question-askers.
% \footnote{To maintain a fair comparison with the subreddit embeddings, we set $d=100$ for the dimensions of the \texttt{Doc2Vec} model. All other model parameters, e.g. negative sampling counts, are set to default.}
The text embedding $e_{t}^{(a)}$ for question-asker $a$ is computed as the average over all prior comments.
% , i.e.
% $e_{t}^{(a)} = \frac{1}{N} \sum_{t \in \text{history}(a)} \mathcal{D}(t)$.

To add the social embedding to the input text, we pass $e^{(a)}$ through a linear layer to match the text dimensionality ($d=768$).
We append a special ``social embedding'' token and the embedding $\hat{e}^{(a)}$ to the end of the text input.
% All readers whose prior posts we were unable to collect are assigned a ``dummy'' embedding of zeroes.

\subsection{Results}
\label{sec:question_generation}

% Having demonstrated consistent differences between question-askers' text (\autoref{sec:compare_reader_groups}), we next test the ability of the question generation model to incorporate reader group information.
We use the models proposed above and a text-only baseline, and train them on the same task of question generation.
We use a sample of our data for training/testing, for a total of 155396 questions for training, 51774 for validation, and 53080 for test.
% data_processing/compute_data_summary_statistics.ipynb#Train/test-split

We use the following metrics to automatically evaluate text quality for target question $q$ and generated question $\hat{q}$: BLEU-1 (single word overlap between $q$ and $\hat{q}$); 
% ROUGE-L (overlap in longest common sub-sequence for $q$ and $\hat{q}$); 
perplexity; 
% Word Mover Distance (mean cosine distance between word embeddings for tokens in $q$ and $\hat{q}$); 
BERT Distance (cosine distance between sentence embeddings for $q$ and $\hat{q}$, via the same DistilBERT system used throughout; ~\citeauthor{sanh2019}~\citeyear{sanh2019}); 
Type/token ratio (among bigrams in $\hat{q}$);
Diversity (\% unique questions among all generated questions $\hat{\mathcal{Q}}$); 
Redundancy (\% generated questions $\hat{\mathcal{Q}}$ that also appear in training data $\mathcal{Q}_{\text{train}}$).
The text overlap metrics like BLEU are important in judging performance even in our open-domain setting, because the models should produce questions that are faithful to the original intent of the question-askers~\cite{wu2021}.
Without measuring overlap, it would be possible for a socially-aware model to generate highly diverse questions that are completely unrelated to the question-asker's intent.

\subsubsection{Aggregate results}

\begin{table*}[t!]
\small
\centering
\begin{tabular}{l p{1.5cm} p{2cm} p{1.5cm} p{2cm} p{2cm} p{1.5cm}}
stat & BLEU-1 $\uparrow$ & BERT Dist. $\downarrow$ & Diversity $\uparrow$ & Type/token $\uparrow$ & Redundancy $\downarrow$ & PPL $\downarrow$ \\ \toprule
Text-only & \textbf{0.159} & \textbf{0.728} & 0.613 & 0.122 & \textbf{0.187} & \textbf{264} \\
Social token & 0.159 & 0.731 & 0.675 & \textbf{0.127} & 0.191 & 271 \\
Social attention & 0.157 & 0.752 & 0.511 & 0.068 & 0.468 & 488 \\
Subreddit embedding & 0.153 & 0.746 & \textbf{0.744} & 0.091 & 0.277 & 657 \\
Text embedding & 0.154 & 0.745 & 0.732 & 0.090 & 0.292 & 609 \\ \bottomrule
\end{tabular}
\caption{Question generation results by model on full test data. $\uparrow$ means higher score is better, $\downarrow$ means lower is better.}
\label{tab:generation_results}
% scripts/models/test_question_generation.sh
% data/reddit_data/text_only_model/test_data_output_scores.tsv
% data/reddit_data/author_text_data/test_data_output_scores.tsv
% data/reddit_data/author_text_data/author_attention_data/author_attention_layer=5_location=encoder_config=attnconcat/test_data_output_scores.tsv
% data/reddit_data/author_text_data/author_subreddit_embed_data/test_data_output_scores.tsv
% data/reddit_data/author_text_data/author_text_embed_data/test_data_output_scores.tsv
\end{table*}

The aggregate results are shown in \autoref{tab:generation_results}.
% \todo{move main results to paper; focus on diversity}
Overall, we see that the simpler socially-aware models (tokens and attention) perform roughly the same as the text-only model via traditional BLEU and BERT Distance metrics.
% In addition, the social-token model outperforms the text-only model in terms of diversity, which shows that being aware of the question-asker may lead to more creative questions with e.g. alternative wording.
The socially-aware model generates questions that have higher overall diversity, but also higher perplexity.
% , which may be explained by overfitting to the specificities of different social groups at the cost of generating less ``on-average'' plausible text.
These results echo prior work in text generation which finds that models which incorporate pragmatic information often produce more diverse text than expected~\cite{schuz2021}.
The higher perplexity can be explained partly by the unconstrained nature of the generation task (e.g., not providing an answer to generate the question; see \autoref{sec:limitations}) as well as the relatively complex nature of most of the questions.
% This  may not be bad depending on the goals of personalized language modeling~\cite{madotto2019}.
% For example...\todo{qualitative examples of high-perplexity questions from reader-aware model}
% Lastly, the embedding models under-perform as compared to the other models, except for diversity.
% The concatenation of social embedding and text input may encourage the model to generate highly unusual questions that are based more on the question-asker's prior experience than on the information provided in the original post.

\begin{table}[t!]
\scriptsize
\centering
\begin{tabular}{p{1cm} p{2.5cm} p{2.5cm}} \toprule
% Subreddit & \texttt{LegalAdvice} & \texttt{AmITheAsshole} & \texttt{PersonalFinance} \\
% Text context & My five year old son is in kindergarten. The teacher let them out of their recess area and did not watch them properly, and my son got lost. & My roommate has been dating someone with a young child. Both the woman and her child are generally annoying. & My partner is turning down salary increases, and she only makes \$60,000 a year in New York. Her main reason is that she's afraid she'll get charged more in payments for her student loans. \\
% Social group & \texttt{LOCATION} (\texttt{US}) & \texttt{EXPERTISE} (\texttt{Novice}) & \texttt{TIME} (\texttt{Fast}) \\ \hline
% Actual question & What is your location? & Have you talked to your roommate? & Does she have access to a 401k that she isn't already maxing? \\
% Text-only & What are your damages? & Have you spoken to your roommate about this? & Is she qualified for the role in california? \\
% Social token & Was this a private school or a government agency? & Do you and your roommate pay rent to the landlord? & Does she have a budget? \\
% Model performance & social token \textgreater \: text-only (BERT Dist.) & text-only \textgreater \: social-token (BLEU) & text-only $\sim$ \: social-token \\ \hline
% \end{tabular}
Subreddit & \texttt{LegalAdvice} & \texttt{AmITheAsshole} \\
Text context & My five year old son is in kindergarten. The teacher let the kids out of their recess area and did not watch them properly, and my son got lost. & My roommate has been dating someone with a young child. Both the woman and her child are generally annoying. \\
Social group & \texttt{LOCATION} (\texttt{US}) & \texttt{EXPERTISE} (\texttt{Novice}) \\ \hline
Actual question & What is your location? & Have you talked to your roommate? \\
Text-only & What are your damages? & Have you spoken to your roommate about this? \\
Social token & Was this a private school or a government agency? & Do you and your roommate pay rent to the landlord? \\
Model performance & social token \textgreater \: text-only (BERT Dist.) & text-only \textgreater \: social-token (BLEU) \\ \hline
\end{tabular}
\caption{Example posts, target questions, and generated questions.}
% models/test_model_response_to_reader_input.ipynb#Test-model-responses:-divisive-posts
\label{tab:example_generated_questions}
\end{table}

\begin{table*}[t!]
\scriptsize
\centering
\begin{tabular}{p{2.5cm} | p{12.5cm} }
\texttt{EXPERTISE} \: \: \: \: \: \: \: \:  \colorbox{blue!50}{\texttt{EXPERT}}, \colorbox{red!50}{\texttt{NOVICE}} &  \colorbox{blue!1.08}{\strut So} \colorbox{blue!16.79}{\strut my} \colorbox{red!50.00}{\strut friend} \colorbox{red!2.68}{\strut is} \colorbox{blue!0.47}{\strut having} \colorbox{blue!2.18}{\strut difficulty} \colorbox{red!8.95}{\strut getting} \colorbox{blue!1.03}{\strut her} \colorbox{blue!5.31}{\strut 15} \colorbox{blue!9.04}{\strut year} \colorbox{blue!1.05}{\strut old} \colorbox{red!16.49}{\strut daughter} \colorbox{blue!2.14}{\strut to} \colorbox{blue!9.50}{\strut school} \colorbox{blue!30.39}{\strut .} \colorbox{blue!27.03}{\strut My} \colorbox{blue!0.00}{\strut friend} \colorbox{red!27.07}{\strut will} \colorbox{red!4.66}{\strut let} \colorbox{blue!4.46}{\strut her} \colorbox{blue!2.53}{\strut off} \colorbox{blue!1.46}{\strut at} \colorbox{blue!5.48}{\strut school} \colorbox{blue!16.08}{\strut ,} \colorbox{blue!10.11}{\strut watch} \colorbox{blue!6.33}{\strut her} \colorbox{blue!8.38}{\strut enter} \colorbox{blue!1.85}{\strut the} \colorbox{blue!11.35}{\strut building} \colorbox{blue!11.54}{\strut ,} \colorbox{blue!19.76}{\strut and} \colorbox{blue!3.88}{\strut then} \colorbox{blue!9.41}{\strut later} \colorbox{blue!0.47}{\strut will} \colorbox{red!42.24}{\strut find} \colorbox{blue!1.88}{\strut her} \colorbox{blue!3.24}{\strut back} \colorbox{blue!3.07}{\strut home} \colorbox{blue!9.01}{\strut during} \colorbox{blue!9.23}{\strut school} \colorbox{blue!9.01}{\strut time} \colorbox{blue!24.88}{\strut .} \\ \hline
\texttt{LOCATION} \: \: \: \: \: \: \: \:  \colorbox{blue!50}{\texttt{NONUS}}, \colorbox{red!50}{\texttt{US}} & \colorbox{blue!1.0}{\strut This} \colorbox{blue!8.53}{\strut happened} \colorbox{red!0.60}{\strut a} \colorbox{blue!44.04}{\strut few} \colorbox{blue!2.37}{\strut days} \colorbox{blue!30.71}{\strut ago} \colorbox{red!8.39}{\strut and} \colorbox{red!6.08}{\strut my} \colorbox{blue!8.13}{\strut friend} \colorbox{red!9.42}{\strut thought} \colorbox{red!13.40}{\strut I} \colorbox{red!0.97}{\strut was} \colorbox{red!1.28}{\strut a} \colorbox{red!0.30}{\strut bit} \colorbox{red!0.26}{\strut rude} \colorbox{red!6.89}{\strut ,} \colorbox{red!3.53}{\strut but} \colorbox{red!24.01}{\strut I} \colorbox{red!6.15}{\strut felt} \colorbox{red!22.47}{\strut I} \colorbox{red!0.00}{\strut was} \colorbox{red!0.52}{\strut totally} \colorbox{red!2.78}{\strut justified} \colorbox{red!17.18}{\strut .} \colorbox{red!3.58}{\strut So} \colorbox{red!3.96}{\strut we} \colorbox{red!50.00}{\strut booked} \colorbox{red!22.04}{\strut tickets} \colorbox{red!4.62}{\strut for} \colorbox{red!1.97}{\strut a} \colorbox{blue!1.93}{\strut nearly} \colorbox{red!0.37}{\strut full} \colorbox{red!11.24}{\strut flight} \colorbox{red!8.29}{\strut and} \colorbox{red!2.46}{\strut the} \colorbox{red!25.66}{\strut only} \colorbox{red!14.97}{\strut row} \colorbox{red!0.42}{\strut with} \colorbox{blue!8.28}{\strut 2} \colorbox{red!13.42}{\strut seats} \colorbox{red!1.90}{\strut beside} \colorbox{blue!19.56}{\strut each} \colorbox{blue!4.86}{\strut other} \colorbox{red!21.22}{\strut had} \colorbox{red!22.82}{\strut somebody} \colorbox{red!15.37}{\strut that} \colorbox{red!12.01}{\strut already} \colorbox{red!17.92}{\strut booked} \colorbox{blue!0.59}{\strut the} \colorbox{red!9.17}{\strut seat...} \\ \hline
\texttt{TIME} \: \: \: \: \: \: \: \: \: \: \: \: \: \: \: \:  \colorbox{blue!50}{\texttt{SLOW}}, \colorbox{red!50}{\texttt{FAST}} & \colorbox{red!28.65}{\strut Folks} \colorbox{red!5.22}{\strut ,} \colorbox{red!7.70}{\strut I} \colorbox{blue!31.28}{\strut am} \colorbox{blue!10.13}{\strut planning} \colorbox{red!5.31}{\strut to} \colorbox{red!6.50}{\strut return} \colorbox{red!5.28}{\strut to} \colorbox{red!17.79}{\strut PCs} \colorbox{red!6.87}{\strut after} \colorbox{red!7.27}{\strut an} \colorbox{red!12.61}{\strut absence} \colorbox{red!5.76}{\strut .} \colorbox{red!5.67}{\strut my} \colorbox{red!15.25}{\strut budget} \colorbox{red!7.69}{\strut is} \colorbox{red!9.87}{\strut about} \colorbox{red!8.17}{\strut 3} \colorbox{red!6.39}{\strut k} \colorbox{red!2.44}{\strut } \colorbox{blue!1.01}{\strut and} \colorbox{red!8.76}{\strut I} \colorbox{red!15.31}{\strut already} \colorbox{red!17.23}{\strut found} \colorbox{red!9.20}{\strut a} \colorbox{red!19.20}{\strut machine} \colorbox{red!7.34}{\strut that} \colorbox{red!11.74}{\strut will} \colorbox{blue!8.87}{\strut be} \colorbox{red!4.23}{\strut around} \colorbox{red!8.57}{\strut 2} \colorbox{red!6.13}{\strut ,} \colorbox{red!9.29}{\strut 5} \colorbox{red!3.62}{\strut k} \colorbox{red!5.07}{\strut .} \colorbox{red!7.47}{\strut So} \colorbox{red!1.93}{\strut right} \colorbox{blue!0.00}{\strut now} \colorbox{red!2.04}{\strut I} \colorbox{red!5.31}{\strut am} \colorbox{red!3.17}{\strut searching} \colorbox{red!5.71}{\strut for} \colorbox{red!12.06}{\strut monitors} \colorbox{red!4.58}{\strut and} \colorbox{red!5.29}{\strut I} \colorbox{blue!14.74}{\strut am} \colorbox{blue!50.00}{\strut looking} \colorbox{red!3.64}{\strut for...}
\end{tabular}
\caption{Ratio of encoder attention generated by social-token model for input conditioned on different social groups. Attention computed via mean over all pairwise scores between tokens.}
\label{tab:attention_distribution_examples}
\end{table*}

% \subsubsection{Example generated questions}

\subsubsection{Qualitative analysis of model output}
\label{sec:qualitative_analysis_model_output}

We first show several examples of generated text (\autoref{tab:example_generated_questions}).
In a legal context (first column), the social-token model correctly predicts that the question-asker will focus on the location of the incident rather than the outcome (text-only model), possibly because a \texttt{US} question-asker may have location-specific advice to provide.
% While the social-token model is incorrect in the second example, in both examples it provides more diverse output than the text-only model.
% For the case of a personal problem (second column), the text-only model correctly predicts that the question-asker will ask about the roommate rather than about the landlord.
% The social-token model may have incorrectly concluded that rent was the main problem, due to many other novice question-askers posing similar finance-related questions when facing similar social problems.

We also use the social-token model to generate attention distributions over the input sequence for different groups.
We input the same text for both reader groups in the same category, changing only the social token appended to the text.
We compute the attention distribution from the first layer of the encoder, compute per-word attention scores via the mean over all heads and all token-pairs, and compute the ratio of attention for each group category.
The distributions for an example post are shown in \autoref{tab:attention_distribution_examples}, and they seem to match our earlier findings with word category differences (\autoref{sec:compare_social_groups}).
For \texttt{LOCATION}, we see that the model prompted with a \texttt{US} token pays more attention to MONEY words (``booked tickets''), while the model prompted with \texttt{NONUS} focuses on time-related words that could be translated to FOCUSPRESENT in the question (``happened,'' ```few days ago'').
For \texttt{Expertise}, the \texttt{NOVICE} social token produces higher attention on social relationships (``friend,'' ``daughter''), and the model with \texttt{EXPERT} input attends to pronouns that could be converted to ``you'' pronouns in a following question (``my'').
For \texttt{TIME}, the model with \texttt{SLOW} input pays attention to DRIVE words (``planning,'' ``looking''), while the model with \texttt{FAST} input pays more attention to personal pronouns (``I'', ``my'').
While we do not perform large-scale annotation of attention distributions, the examples shown here complement the generated text and reveal potential concepts that the model has learned to associate with different social groups.

\subsubsection{Divisive posts}
\label{sec:divisive_posts}
% When considering how questions vary across different reader groups, it is important to identify posts where reader groups ask very different questions.
Socially-aware question generation should perform well in cases where different social groups have divergent opinions, e.g. where experts disagree with novices.
We now test the models' ability to predict divisive questions.
% We quantify this notion by testing on pairs of semantically-divergent questions from different reader groups.
For post $p$, question $q_{1}$ written by an author of group 1, and question $q_{2}$ written by an author of group 2, we define $\text{sim}(q_{1}, q_{2})$ based on the cosine similarity of the latent representations of the questions, generated by DistilBERT as before~\cite{sanh2019}.
We 
% identify  posts with pairs of question-askers of different groups, then 
label as ``divisive'' all pairs of questions that have a similarity score in the lowest $n^{\text{th}}$ percentiles.
% We also test static word embeddings to measure question divisiveness and find similar results (see \autoref{sec:divisive_posts_word_embeddings}).
% for a total of 1073 posts and 2146 questions.\footnote{The low numbers are due to incomplete coverage of reader data for all questions and the fact that many posts don't attract more than one question associated with a particular reader group, after data filtering (\autoref{sec:data}).}
% We label these data instances as \emph{divisive} posts, because they attract divergent questions from different social groups.
We show examples of divisive posts in \autoref{sec:divisive_question_examples}.
% One example post asks for advice about paying for a car, and a \texttt{Novice} reader asks ``Are you above water on the car?'' while a \texttt{Expert} reader asks ``Have you been applying for jobs all day?''
% scripts/data_processing/compare_posts_different_reader_responses.ipynb#Compare-posts-with-different-reader-responses
% Ideally, a socially-aware model should generate both types of questions. 
% If a post writer wants advice from fellow novices, they might see the generated \texttt{Novice} question and then add information about car payments to their post to help other novices in the audience provide their final advice.

% to accurately anticipate such questions is critical to helping the post author understand what different question-askers need to know about to provide good advice.

The results of the question prediction task on divisive posts are shown in \autoref{tab:divisive_post_generation_results}.
The social-token model slightly outperforms the text-only baseline for questions that are highly dissimilar (i.e. less similar than 90\% and 95\% of the question pairs), and all socially-aware models tend to do better in diversity.
This suggests that the social-token model may pick up information specific to the different social groups that is required to anticipate how the question-askers approach potentially subjective posts.
We also note the unusually high perplexity across all models, which may indicate that socially-specific questions are complicated and far from ``normal'' questions. 
% , e.g., a \texttt{Novice} asker requesting details on payment for the car as compared to a \texttt{Expert} asker curious about underlying financial issues.

\begin{table}[t!]
\small
\centering
% \begin{tabular}{>{\raggedright\arraybackslash}p{3cm} r r r}
% & BLEU-1 & ROUGE-L & PPL \\ \toprule
% \multicolumn{4}{c}{sim($q_{1}$, $q_{2}$) $\leq$ 5\% (N=1074)} \\ \hline
% Text-only & 0.137 & 0.086 & 383 \\
% Social token & \textbf{0.142} & 0.093 & \textbf{359} \\
% Social attention & 0.130 & 0.082 & 601 \\
% Subreddit embedding & 0.137 & \textbf{0.094} & 945 \\
% Text embedding & 0.137 & 0.088 & 623 \\ \hline
% \multicolumn{4}{c}{sim($q_{1}$, $q_{2}$) $\leq$ 10\% (N=2146)} \\ \hline
% Text-only & 0.160 & 0.110 & \textbf{325} \\
% Social token & \textbf{0.164} & \textbf{0.119} & 327 \\
% Social attention & 0.155 & 0.113 & 547 \\
% Subreddit embedding & 0.148 & 0.104 & 1048 \\
% Text embedding & 0.150 & 0.099 & 617 \\ \bottomrule
% \end{tabular}
\begin{tabular}{p{2cm} r r r r}
Model & BLEU-1 & \multicolumn{1}{l}{Div.} & \multicolumn{1}{l}{Red.} & PPL \\ \hline
\multicolumn{5}{c}{sim($q_{1}$, $q_{2}$) $\leq$ 5\% (N=1074)} \\ \hline
Text-only & 0.137 & 0.688 & 0.222 & 383 \\
Social token & \textbf{0.142} & 0.771 & \textbf{0.208} & \textbf{359} \\
Social attention & 0.130 & \textbf{0.875} & 0.479 & 601 \\
Subreddit emb. & 0.137 & 0.854 & 0.292 & 945 \\
Text emb. & 0.137 & 0.840 & 0.375 & 623 \\ \hline
\multicolumn{5}{c}{sim($q_{1}$, $q_{2}$) $\leq$ 10\% (N=2146)} \\ \hline
Text-only & 0.160 & 0.699 & \textbf{0.232} & \textbf{325} \\
Social token & \textbf{0.164} & 0.781 & 0.235 & 327 \\
Social attention & 0.155 & 0.798 & 0.500 & 547 \\
Subreddit emb. & 0.148 & 0.864 & 0.308 & 1048 \\
Text emb. & 0.150 & \textbf{0.868} & 0.348 & 617 \\ \hline
\end{tabular}
\caption{Question generation results for divisive posts.}
\label{tab:divisive_post_generation_results}
% raw data: data/reddit_data/paired_question_low_sim_simpct=10_data.gz; data/reddit_data/paired_question_low_sim_simpct=5_data.gz
% spreadsheet: https://docs.google.com/spreadsheets/d/1C2sX2bmTVbKwkzGaI_nwRmFJqEFhC42qE16-EryIX24/edit#gid=2014324046
\end{table}

\subsubsection{Group-specific questions}
\label{sec:generation_social_specific_questions}

We investigate another desired property of socially-aware models, the ability to predict questions that are strongly associated with a particular group.
Post writers would benefit from such questions, e.g. technical questions from ``expert'' askers, because these questions would help the post writer preempt specific and unexpected information needs from that group.
% A socially-aware question-generation model should provide insight to the post writer about questions that are highly specific to a particular group
% Ideally, a reader-aware question-generation model would perform well on questions that are highly associated with a particular reader group, e.g. a prototypical \texttt{Expert} question.
We subset the data to all questions $q$ with question-asker $a$ that the trained social group classifiers assign to the group $g_{a}$ with high confidence ($P \geq 95\%$) (see \autoref{sec:social_group_classification} for classifier details).
% If $q$ is written by an \texttt{Expert} question-asker and $P(r = \texttt{Expert} | q, p) \geq 95\%$, we keep $q$.

\begin{figure}[t!]
    \centering
    \includegraphics[width=\columnwidth]{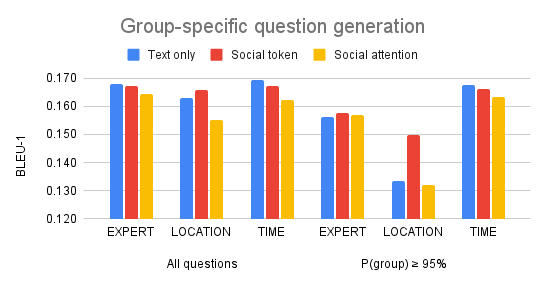}
    \caption{BLEU-1 scores for question generation, on (1) full data and (2) subset of data with high group-specific probability (determined by classifier).}
    \label{fig:social_specific_question_generation}
    % raw data: data/reddit_data/group_classification_model/question_post_data/reader_group_cutoff_pct=95.gz
% spreadsheet: https://docs.google.com/spreadsheets/d/1C2sX2bmTVbKwkzGaI_nwRmFJqEFhC42qE16-EryIX24/edit#gid=278356316
\end{figure}

% \todo{shorter}
We report the results for this data subset in \autoref{fig:social_specific_question_generation}.
The relative performance of the socially-aware models increases when only considering data with highly group-specific questions.
This is particularly apparent for the \texttt{LOCATION} group category, illustrated by the following example.
% for which $P(r | q, p) \geq 95\%$.
% Furthermore, performance generally drops when switching from all questions to reader-specific questions, but the drop is less severe for the reader-aware models.
% This reinforces our prior finding about posts that attract highly different questions: the socially-aware models are more suited for questions with highly distinct perspectives from different social groups.
% We find the biggest performance difference in socially-specific questions for the \texttt{LOCATION} questions, which is in line with prior work that found gains in personalization for generation conditioned on demographics such as location~\cite{welch2020}.
In our data, a socially-specific question was written by a \texttt{non-US} question-asker in \texttt{LegalAdvice} in response to a post about a mailing problem: ``Have you sent a change of address notice to the post office?''
In this situation, the social-token model generated the question ``Did you give them your current address?''
The social-token model seems to have identified a concern that a non-US question-asker might be more likely to focus on (e.g. due to moving frequently) than a US question-asker.

% \subsection{Error analysis}

% We assess the relative performance of the reader-aware models across different conditions, to assess their potential value for downstream applications.

\subsection{Human evaluation}
\label{sec:generation_annotation}

% It is not enough to verify that the model can predict text with relatively high accuracy.
% We next determine if the model generates text that seems relevant to a particular reader group based on human evaluation~\cite{nema2018}.
To corroborate the generation results about divisive questions, we collect human annotations about: (1) question quality; and (2) guessing the social group based on the generated question (see \autoref{sec:generation_annotation_social_group} for (2)).
% The task (2) tests whether the reader-aware model's questions accurately reflect the reader group's perceived identity, e.g., whether question $q$ reflects background knowledge that an \texttt{Expert} reader would be more likely to have than a \texttt{Novice} reader.
% % We collect annotations from M evaluators recruited from the original subreddits...\todo{dazzling results}

We use the text-only model and the social-token model to generate questions from a sample of the test data, as follows.
% For each subreddit $s$ and reader group category $G$, we sample 10 posts that have a question from a reader of a particular group: e.g., to represent the \texttt{EXPERTISE} group reader, we include a post $p$ with $q_{1}$ from either an \texttt{Expert} reader or a \texttt{Novice} reader.
For each subreddit $s$ and social group category $\mathcal{G}$, we sample up to $N=10$ posts that have divisive questions from groups $g_{1}$ and $g_{2}$
% , where $\text{sim}(q_{1}, q_{2}) \leq 10\%$ according to the overall distribution of $\text{sim}(q_{1}, q_{2})$ within the subreddit and group category 
where the similarity is below the $10^{\text{th}}$ percentile (\autoref{sec:divisive_posts}).\footnote{Some combinations of subreddits and reader groups have fewer than 10 posts, due to the data sampling strategy.}
We then generate a single question for the post from the text-only model 
% ($\hat{q}_{t}$) 
and two questions from the social-token model,
% ($\hat{q}_{1,r}$, $\hat{q}_{2,r}$), 
one for each social group in the category (e.g., for \texttt{EXPERTISE}, $q_{1}$ for \texttt{Expert} and $q_{2}$ for \texttt{Novice}).
We provide details of annotation in \autoref{sec:annotation_details}.

\begin{table}[t!]
\centering
\small
\begin{tabular}{l l l l}
Text type & \textbf{A} & \textbf{R} & \textbf{U} \\ \toprule
\multicolumn{4}{c}{\textbf{Overall}} \\ \hline
Ground-truth & 3.83 & 3.59 & 3.92 \\
Text-only & \underline{3.84} & \underline{3.68}* & \underline{3.96}* \\
Social-token & 3.80 & 3.35 & 3.73 \\ \hline
\multicolumn{4}{c}{\textbf{Social group}} \\ \hline
\multicolumn{4}{c}{\texttt{EXPERTISE}} \\
Ground-truth & 3.89 & 3.68 & 3.85 \\
Text-only & \underline{3.81} & \underline{3.62}* & \underline{3.91}* \\
Social-token & 3.61 & 2.99 & 3.49 \\
\multicolumn{4}{c}{\texttt{LOCATION}} \\
Ground-truth & 4.06 & 3.58 & 4.20 \\
Text-only & 4.01 & \underline{3.69} & 4.05 \\
Social-token & \underline{4.20} & 3.63 & \underline{4.19} \\
\multicolumn{4}{c}{\texttt{TIME}} \\
Ground-truth & 3.64 & 3.52 & 3.83 \\
Text-only & \underline{3.77} & \underline{3.74} & \underline{3.95}* \\
Social-token & 3.74 & 3.53 & 3.70 \\ \hline
\multicolumn{4}{c}{\textbf{Subreddit}} \\ \hline
\multicolumn{4}{c}{\texttt{Advice}} \\
Ground-truth & 3.75 & 3.63 & 3.91 \\
Text-only & 3.32 & \underline{3.29} & 3.57 \\
Social-token & \underline{3.49} & 3.15 & \underline{3.67} \\
\multicolumn{4}{c}{\texttt{AmItheAsshole}} \\
Ground-truth & 3.79 & 3.58 & 4.01 \\
Text-only & 3.74 & \underline{3.61} & \underline{3.89} \\ 
Social-token & \underline{3.82} & 3.39 & 3.69 \\ 
\multicolumn{4}{c}{\texttt{LegalAdvice}} \\
Ground-truth & 4.18 & 3.88 & 4.47 \\
Text-only & \underline{3.95} & \underline{3.60} & \underline{4.19}* \\
Social-token & 3.86 & 3.23 & 3.81 \\ 
\multicolumn{4}{c}{\texttt{PCMasterRace}} \\
Ground-truth & 3.72 & 3.44 & 3.62 \\
Text-only & \underline{4.20} & \underline{4.07}* & \underline{4.16} \\
Social-token & 3.98 & 3.39 & 3.84 \\
\multicolumn{4}{c}{\texttt{PersonalFinance}} \\
Ground-truth & 3.72 & 3.43 & 3.58 \\
Text-only & \underline{4.04} & \underline{3.89} & \underline{4.01}* \\ 
Social-token & 3.87 & 3.56 & 3.70 \\ \hline
\end{tabular}
\caption{Human annotation scores for question quality, including \textbf{A}nswerable, \textbf{R}elevant, \textbf{U}nderstandable (scale 1-5). 
* indicates that the score is greater than the scores from the other model type with $p>0.05$ (Wilcoxon test). 
\underline{Underline} indicates best generation model.
}
\label{tab:annotation_question_quality_scores}
% scripts/models/collect_generated_text_eval_data.ipynb#Prolific-test:-official-round-2
\end{table}

We show the results in \autoref{tab:annotation_question_quality_scores}.
The annotators in aggregate preferred the questions from the text-only model over the social-token model.
However, the social-token model questions were perceived as more answerable and understandable for questions generated using \texttt{LOCATION} information, which aligns with prior results (\autoref{sec:generation_social_specific_questions}).
The social-token model is also perceived as more answerable and understandable in the context of \texttt{Advice}, which makes sense considering that the social-token model has more diverse output that may suit the broad domain of general-advice posts.
% might be attributed to the more generic nature of the subreddit's posts that lend to different interpretations from different social groups.
% , and therefore ask different questions.

\begin{table}[t!]
\scriptsize
\centering
\begin{tabular}{p{1.25cm} p{2.5cm} p{2.5cm}} \toprule
Subreddit & \texttt{LegalAdvice} & \texttt{Advice} \\
Text context & My mother lost \$50000 on an online dating site to a scam. If something happened to her, would I be on the hook for this? & I want to break up with my girlfriend but: number 1 I don't want to hurt her, number 2 I don't know if I can manage on my own, number 3 I don't always believe in myself, and if I lose my job I'll be homeless. \\
Social group & \texttt{EXPERTISE} (\texttt{Expert}) & \texttt{LOCATION} (\texttt{non-US}) \\ \hline
Actual question & Has your mother contacted the police? (Understandable=4.67) & Have you tried talking to her? (Answerable=5) \\
Text-only model & How did the scammer get the info from your Mom? (Understandable=4.33) & Number 2 doesn’t even sound like a good idea, have you tried number 3? (Answerable=2.33) \\
Social token model & Are you on the hook for what? (Understandable=2.67) & Why do you think you'll be homeless? (Answerable=5) \\ \bottomrule
 &  & 
\end{tabular}
\caption{Example questions with human evaluation scores.}
% models/collect_generated_text_eval_data.ipynb#Show-example-generated-questions-with-scores
\label{tab:example_human_evaluation_questions}
\end{table}

We show example generated and actual questions with their human evaluation ratings in \autoref{tab:example_human_evaluation_questions}.
In the first example, the text-only model addresses an important missing gap in original post (how the scammer got information), while the social-token model seems to focus too much on details (``on the hook'') which leads to a less understandable question.
In the second example, the social-token model addresses missing information that may be more salient to a non-US question-asker who wants to know more about homelessness (possibly less salient to a US question-asker), while the text-only model produces a question that is not answerable due to a misunderstanding of the original post (focus on the text rather than the writer).
Note that this type of question is not marked by a surface-level feature such as regional style, but rather a deeper focus on cause and effect, which suggests that the model has learned more fundamental differences about the nature of \texttt{LOCATION} as a social group.

\section{Conclusion}

\subsection{Discussion}
% \todo{shorter; focus on answering analysis questions}
This study evaluated the incorporation of social information into question generation, to help writers understand the information needs of different people.
% This suggests that future work in question generation should focus on other types of ``concrete'' demographic variables, because demographics may reflect a person's everyday experiences and information needs better than e.g. domain expertise.
We found that social groups related to expertise, time, and location can all be differentiated based on the questions that they ask. 
In generation, the discrete social representations outperformed continuous representations,
% , as the social-token model consistently outperformed the more complicated social-representation models.
and the social-token model outperformed the baseline when the questions are divisive.
In human evaluation, the social-token models produced better output for the \texttt{LOCATION} group, implying a more clear definition versus other social groups.
% , which indicates that a question-asker's location is more informative as compared to \texttt{EXPERTISE} and \texttt{TIME}.

Future research in question generation should focus on divisive questions as the main area of improvement.
Researchers may also consider ensemble models~\cite{liu2021} that use a text-only generator with less subjective input text (e.g., in technical settings), and a social-token generator in more divisive settings.
% In the case of the \texttt{EXPERTISE} group, this study's data set can be analyzed further to provide more flexible representations of expertise for question generation, e.g. treating expertise as a spectrum rather than a binary.
For future evaluation, socially-aware question generation may benefit other contexts such as journalism, medicine, and public policy, where people are likely to have differing information needs based on their background experience~\cite{assmann2017}.
No matter the case, writers will always benefit from knowing in advance what information their audience will need.

\subsection{Limitations}
\label{sec:limitations}

The primary limitations of this work relate to the definition of ``social group,'' which may have contributed to the minimal gains by the social token model.
This work focused on generic social groups that can be extended to other domains, which may leave out domain-specific social groups (e.g., socioeconomic status).
The social groups may not mean the same thing in different domains: an \texttt{EXPERT} question-asker in the legal domain may be a professional lawyer, while in personal advice the average \texttt{EXPERT} may lack professional experience.
Most notably, the social groups used in this work were not validated by any annotators or by the question-askers.
This especially matters for the \texttt{EXPERTISE} category, considering the subjective status of expertise within online communities~\cite{johnson2001}.
To accurately identify non-obvious social groups, researchers should ask domain experts to label a small set of user data as gold labels, and then compare the automated labels against this gold standard set.

In terms of the task, this work focuses on unconstrained question generation, i.e. we do not use answers~\cite{dong2019} or intentions~\cite{cao2021} to guide generation.
The results presented in this work represent a lower bound on performance, which includes unusually high perplexity (\autoref{tab:generation_results}) and sometimes unexpected topic choices (\autoref{tab:example_generated_questions}).
This problem is compounded by the fact that social group information may not always be useful e.g. for non-divisive questions, and therefore such social guidance may simply confuse the model.
Future work would collect both questions and answers, or at least question type labels, to provide consistent guidance for socially-aware question generation.

\section{Ethics statement}

We acknowledge that text generation is an ethically fraught application of NLP that can be used to manipulate public opinion~\cite{zellers2020} and reinforce negative stereotypes~\cite{bender2021}.
Our models could be modified to generate abusive or factually misleading questions, which we do not endorse.
Furthermore, our models may accidentally memorize private information from the training data.
We intend for our work to benefit people who share information about themselves for the purpose of gaining feedback from peers.

All data used in this project was publicly available via the Pushshift API~\cite{baumgartner2020}.
In our final release we will not release any data with personally identifiable information (e.g., \texttt{LOCATION} data), in order to protect the original authors.
This is not ideal considering that \texttt{LOCATION} seemed to be the most useful input to the model, but the remaining social attributes may prove useful for future researchers who want to test other definitions of ``divisive'' questions, e.g. positive versus negative valence.
% This includes identifying posts where people ask divergent questions in terms of positive vs. negative valence.
Furthermore, we do not claim that we have the perfect definitions of the social groups that we attempted to identify in our study, and it is possible that a Reddit user who finds themselves labeled as e.g. an ``expert'' would disagree.
We encourage future researchers to compare their own definition of the various social groups against our own labels, e.g. a different definition of ``expertise.''

\section*{Acknowledgments}
This material is based in part on work supported by the John Templeton Foundation (grant \#61156) and by the Michigan Institute for Data Science (MIDAS). Any opinions, findings, conclusions, or recommendations in this material are those of the authors and do not necessarily reflect the views of the John Templeton Foundation or MIDAS.
We thank members of the LIT Lab, including Artem Abzaliev and Aylin Gunal, for their help piloting the human annotation task, and for providing feedback on early results.
We also thank the annotators who identified valid questions as part of the data filtering process.

\bibliography{main}
\bibliographystyle{acl_natbib}

\appendix

\section{Data: question filtering}
\label{sec:data_question_filtering}

Initial analysis revealed that some questions were either irrelevant to the post (e.g., ``what about X'' where X is unrelated to the post topic) or did not actually seek more information from the original post (e.g., rhetorical questions).
% These conditions make a question invalid, according to our assumptions about rational speakers who seek new information from one another.
To address this, we sampled 100 questions from each subreddit in the data along with the parent post, and we collected binary annotations for relevance (``question is relevant'') and information-seeking (``question asks for more information'') from three annotators, who are undergraduate students and native English speakers.
We provided instructions and a sample of 20 questions labeled by one of the authors as training data for the annotators.
% , and one of the authors met with the annotators after training to resolve disagreements.
On the full data, the annotators achieved fair agreement on question relevance ($\kappa=0.56$) and on whether questions are information-seeking ($\kappa=0.62$).
% mainly due to class imbalance (about 83\% ``relevant'' annotations and 17\% ``irrelevant'' annotations)~\cite{feinstein1990}.

% The annotations showed reasonable agreement according to Krippendorff's alpha (X indicates ``good'' quality agreements CITE: prior work).
After annotation, we removed all instances of disagreement among annotators to yield questions with perfect agreement for relevance (76\% perfect-agreement) and information-seeking (80\%).
In the perfect-agreement data, the majority of questions (94\%) were marked as relevant by both annotators, which makes sense considering that the advice forums generally attract good-faith responses from commenters.
% most people who take time to ask a follow-up question in the advice subreddits studied are likely acting in good faith.
We therefore chose to not filter questions based on potential relevance.
% Using this data as a bootstrap, we automatically identified relevant and information-seeking questions using two heuristics.
% For relevance, we computed the token-level overlap between each post $p$ and the associated question $q$ as the maximum overlap score over every sentence $s$ and $q$:
% $$\text{overlap}(p,q) = \underset{s \in p}{\text{max}} \: \: \text{BLEU-1}(s, q)$$
% We marked a question as relevant if the overlap score was in the interval [0.1, 0.5], which maximizes precision (95\%) without excessively hurting recall (48\%).
To filter information-seeking questions, we trained a simple bag-of-words classifier on the annotated data (binary 1/0; based on questions with perfect annotator agreement).\footnote{We restricted the vocabulary to the 50 most frequent words, minus stop-words, to avoid overfitting. Initial tests with SVM, logistic regression, and random forest models revealed that the random forest model performed the best, which we used for the final classification model.}
The annotated data were split into 10 folds for training and testing, and the model achieved 87.5\% mean F1 score, which is reasonable for ``noisy'' user-generated text.
% and which improves our confidence in the annotations.
We applied the classifier to the full dataset and removed questions for which the classifier's probability was below 50\%.

% After filtering the original data using the heuristics for question relevance and information-seeking, we arrived at 730,620 valid questions total for training and testing.

\section{Defining social groups}

\subsection{Social embeddings: topically-related subreddits}
\label{sec:model_topical_subreddits}

\begin{table}[t!]
\centering
\small
\begin{tabular}{l >{\raggedright\arraybackslash}p{4cm}}
\toprule
Subreddit & Neighbors \\
\midrule
\texttt{Advice} & \texttt{answers, ask, askdocs, dating\_advice, getdisciplined, mentalhealth, needadvice, socialskills, tipofmytongue} \\ \hline
\texttt{AmItheAsshole} & \texttt{askdocs, isitbullshit, tooafraidtoask} \\ \hline
\texttt{LegalAdvice} & \texttt{askhr, bestoflegaladvice, insurance, landlord, lawschool, legaladviceuk, scams} \\ \hline
\texttt{PCMasterRace} & \texttt{bapcsalescanada, buildmeapc, linuxmasterrace, monitors, overclocking, pcgaming, suggestalaptop, watercooling} \\ \hline
\texttt{PersonalFinance} & \texttt{accounting, askcarsales, churning, creditcards, financialindependence, financialplanning, investing, realestate, smallbusiness, studentloans, tax, whatcarshouldibuy, yna} \\
\bottomrule
\end{tabular}
 \caption{Filtered neighbor subreddits for advice-related subreddits.}
\label{tab:subreddit_neighbors}
% data/reddit_data/author_data/advice_subreddit_neighbors.tsv
\end{table}

In our discrete-representation models, the criterion for defining a question-asker for post $p$ in subreddit $s$ as an \texttt{Expert} or \texttt{Novice} is whether they have previously written comments in $s$ or in a topically similar subreddit.

We find similar subreddits for each target subreddit $s$ by (1) computing the top-20 nearest neighbors for subreddit $s$ in subreddit embedding space (see \autoref{sec:social_embeddings}) and (2) manually filtering unrelated subreddits.
We report the related subreddits in \autoref{tab:subreddit_neighbors}.

\subsection{Validating group differences: classification}
\label{sec:social_group_classification}

To verify the differences in question content observed in \autoref{sec:compare_social_groups}, we train a single-layer neural network to classify social groups, using a latent semantic representation of the question-asker's question $q$ and the related post $p$ generated by the DistilBERT transformer model~\cite{sanh2019}.
The embedding for the question and the post are each converted to $d=100$ dimensions via PCA for regularization, and then concatenated.
% We trained the model to convergence using Adam optimization.
We train a separate model for each subreddit, and we up-sample data from the minority class.
% For each subreddit, we up-sample data to balance the class distributions, and we train a separate model for each subreddit.
% , under the assumption that the reader groups would behave differently depending on the topic (e.g. an expert in finance would ask different questions compared to an expert in computers).%We experimented with training a transformer model for sequence classification to predict reader groups but found sub-optimal performance for some of the groups. 

\begin{table}[]
    \centering
    \small
    \begin{tabular}{p{1.7cm} p{2.0cm} p{1.8cm}}
        Features & Social group & Accuracy \\ \hline
        \multirow{3}{1.7cm}{Question text} & \texttt{EXPERTISE} & 70.1 ($\pm$ 2.5)  \\
        & \texttt{TIME} & 81.6 ($\pm$ 7.5) \\
        & \texttt{LOCATION} & 75.4 ($\pm$ 2.8) \\ \midrule
        \multirow{3}{1.7cm}{Post + question text} & \texttt{EXPERTISE} & 73.5 ($\pm$ 6.4) \\
        & \texttt{TIME} & 83.1 ($\pm$ 8.3) \\
        & \texttt{LOCATION} & 66.3 ($\pm$ 10.2) \\ \bottomrule
        % data/reddit_data/group_classification_model/question_data/combined_score_data.tsv
        % data/reddit_data/group_classification_model/question_post_data/combined_score_data.tsv
        % old results: one model for all communities
        % \multirow{3}{2.5cm}{Question text} & \texttt{EXPERT} & 66.9 ($\pm$ 0.4) \\
        % & \texttt{TIME} & 88.9 ($\pm$ 0.3) \\
        % & \texttt{LOCATION} & 65.8 ($\pm$ 1.4) \\ \midrule
        % \multirow{3}{2.5cm}{Post + question text} & \texttt{EXPERT} & 80.4 ($\pm$ 0.4) \\
        % & \texttt{TIME} & 91.4 ($\pm$ 0.4) \\
        % & \texttt{LOCATION} & 65.9 ($\pm$ 1.4) \\ \midrule
        
    \end{tabular}
    \caption{Social group prediction accuracy (mean, standard deviation measured across subreddits). 
    % The models were all single-layer linear neural networks, trained on the DistilBERT representation of questions and (optionally) posts.
    }
    \label{tab:reader_group_prediction_scores}
    % scripts/data_processing/compare_reader_group_questions.ipynb#Classification-via-semantic-representations
\end{table}

% we train a bag-of-words classifier\footnote{We used an SVM after preliminary tests showed higher performance across conditions than other classifiers. We restricted the vocabulary to the top-500 features after filtering stop-words, to prevent overfitting.} to predict the reader group that produced a given question.

We report mean accuracy over all subreddits in \autoref{tab:reader_group_prediction_scores}.
The models consistently outperform the random baseline across all group categories tested, which suggests a clear difference between social group members.
The models trained on the combined post and question text generally help prediction improve over the question text alone, which supports the hypothesis that a question-asker's background is reflected in both the question they ask and the context in which the question is asked.
Therefore, generating group-specific questions requires understanding how the question relates to the original post content, in addition to the question writing style.
We find an unusually high performance for \texttt{TIME}, which may be due to a more consistent writing style among \texttt{Fast} question-askers.

\section{Results: question generation}

We report here the results of additional tests to evaluate the relative utility of the socially-aware models with respect to different types of question-post scenarios.

% \subsection{Aggregate results}

% We present the full results in \autoref{tab:generation_results}.

\subsection{Performance by question type}

\begin{figure*}[t!]
    \centering
    \includegraphics[width=0.7\textwidth]{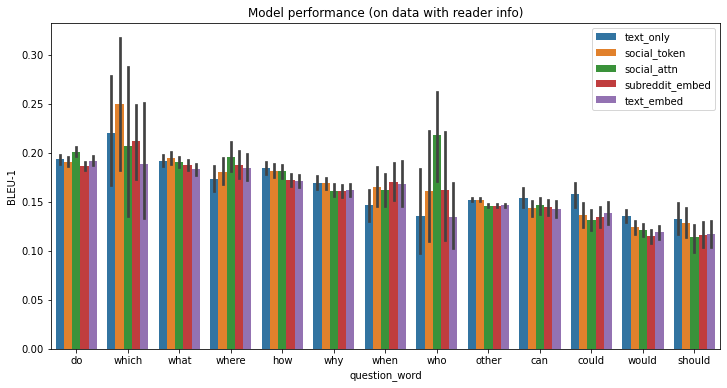}
    \caption{Model performance by question type.}
    \label{fig:model_question_types}
    % scripts/models/error_analysis.ipynb#Compare-questions:-type
\end{figure*}

First, we assess the relative performance of different question generation models according to the type of question asked.
Questions are categorized based on the root question word, e.g. ``who,'' ``what,'' ``when.''\footnote{We use the dependency parser from \texttt{spacy}~\cite{honnibal2015} to identify root question words based on their dependency to the \texttt{root} verb of the question (e.g. \texttt{advmod} for ``where'' in ``where do you live?'').}
We compare the BLEU-1 scores of all question generation models on the specified questions, restricting to questions asked by question-askers who could be assigned to at least one social group or an embedding.

The results are shown in \autoref{fig:model_question_types}.
In contrast to the aggregate results, the social-attention model outperforms the text-only baseline for ``do,'' ``where,'' and ``who'' questions.
All socially-aware models outperform the text-only model for ``when'' questions.
These questions may reflect more of a focus on concrete details such as locations, times, and people mentioned by the original poster, and therefore the socially-aware models may generally identify differences among question-askers in terms of the details requested.
The text-only model outperforms the socially-aware models for questions that are potentially more subjective, including ``can,'' ``could,'' ``would,'' and ``should'' questions.
These more subjective questions may require the models to focus more precisely on the original post (e.g. a ``would'' question to pose a hypothetical concern about the post author's situation), and therefore such questions may be less dependent on question-asker identity.

% The baseline text-only model does the best on less common question types, including those that begin with ``where'' and ``when'' (i.e. focusing more on details of the post itself rather than subjective interpretations by the reader).
% We find that the reader-attention model does surprisingly well on questions that begin with ``what,'' ``which,'' ``how,'' and ``who,'' which suggests that reader background may be important for questions that relate to more subjective or social interpretations of the original post.
% The reader embedding models both perform well on ``do'' questions, which may relate to the fact that such questions are more direct and potentially more related to the question asker's prior experience rather than original post author.

\subsection{Post similarity}

A helpful question should be related to the original post, but should not be so similar that it requests information that the post has already provided.
We therefore assess the tendency of the models to generate semantically related questions for the given posts.
We compute the similarity between each generated question $q$ and the associated post $p$ using the maximum cosine similarity between the sentence embedding for $q$ and each sentence $s$ in $p$.
The sentence embeddings are generated using the DistilBERT model~\cite{sanh2019}.
% \footnote{While we acknowledge that metrics like \texttt{QBLEU} could also be used here to evaluate question generation specifically using token overlap between the question and related document~\cite{nema2018}, we use latent semantic representations instead to identify semantic similarities that may not appear obvious from the text tokens alone. E.g. if a question in \texttt{PCMasterRace} mentions ``motherboard'' in response to a post, this may cause the question to score poorly with a typical overlap metric despite the semantic relevance of the question to the post.}

\begin{figure}[t!]
    \centering
    \includegraphics[width=\columnwidth]{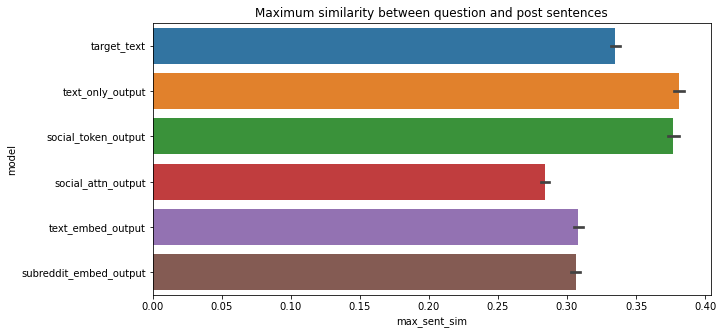}
    \caption{Maximum semantic similarity between questions and sentence from original post.}
    \label{fig:max_sent_sim_questions}
    % scripts/models/error_analysis.ipynb#Post-overlap:-semantics
\end{figure}

The results in \autoref{fig:max_sent_sim_questions} show that the best overall models, text-only and social-token, generate questions that are more similar to the original post than expected (cf. ``target text'' i.e. ground-truth).
The other socially-aware models show a significantly lower similarity, implying that their generated questions address \emph{new information} about the post that is not mentioned in the post itself.
For example, in response to a \texttt{r/Advice} post about self-improvement (``I just need some tips on maybe motivating myself''), the model with social text embeddings asks ``What do you want to do with your life?''
The generated question is less semantically similar to the original post than the target question (``Have you talked to a doctor about this?'') but addresses an underlying personal issue for the post author that only a particularly thoughtful question-asker would uncover.

\subsection{Divisive questions: examples}
\label{sec:divisive_question_examples}

\begin{table*}[t!]
\centering
\small
\begin{tabular}{l p{4cm} p{4cm} p{4cm}} \toprule
Subreddit & \texttt{PersonalFinance} & \texttt{LegalAdvice} & \texttt{AmITheAsshole} \\
Text context & I need help figuring out what's the best next step. I have \$1200 saved for car payments but I have no idea after that. & Last month I got a letter from a law firm representing someone that I owe a debt to. Two years ago I couldn't continue to make payments to the creditor and almost went bankrupt. & My younger brother is autistic. He can function and he has a job (janitor), hangs out with his friends but he can't live on his own. \\
Social group & \texttt{EXPERTISE} & \texttt{TIME} & \texttt{LOCATION} \\ \midrule
Group 1 & (\texttt{Expert}) Have you been applying for jobs all day? & (\texttt{Slow}) Have you asked what they are willing to settle for? & (\texttt{US}) What if down the road you had to re-locate for work or your wife's work? \\
Group 2 & (\texttt{Novice}) Are you above water on the car? & (\texttt{Fast}) Do you actually intend on filing bankruptcy? & (\texttt{non-US}) How disabled is your brother? \\
Question similarity & 0.209 & 0.256 & 0.190 \\ \bottomrule
\end{tabular}
\caption{Example divisive questions for different social groups.}
\label{tab:example_divisive_questions}
\end{table*}

We provide examples of divisive posts in \autoref{tab:example_divisive_questions} (\autoref{sec:divisive_posts}).
For \texttt{TIME}, the \texttt{Slow} question-asker seems to target a more complicated and underlying issue around the debt problem, while the \texttt{Fast} question-seeker clarifies a basic detail about the case.
For \texttt{LOCATION}, the question from the \texttt{US} asker focuses on adapting to work needs, while the \texttt{non-US} question addresses the writer's brother and his medical situation.
In all cases, we can see that these kinds of questions are more likely to be anticipated by a generation model that produces more diverse output.

\subsection{Divisive posts: word embeddings}
\label{sec:divisive_posts_word_embeddings}

In \autoref{sec:divisive_posts}, we identified questions as ``divisive'' based on low similarity between the latent representations of the questions, as generated by a sentence encoder. 
We also experiment with determining divisiveness based on static word embeddings.
We leverage a set of word embeddings trained with the FastText algorithm~\cite{joulin2017}, and we convert each questions to a latent representation using the average over all embeddings for the tokens in the question.
We then compute paired question similarity as before, with cosine similarity.
The questions from the sentence embeddings and those from the static word embeddings have a high degree of overlap: setting the similarity threshold below 5\% yields an overlap of 23.7\%, and a similarity threshold below 10\% yields an overlap of 45.3\%.
Next, we test the correlation between the sentence embedding similarity and the word embedding similarity and find a high amount of correlation ($R=0.98$, $p<0.001$).
We conclude that labeling divisive questions using word embedding similarity rather than sentence embedding similarity would yield similar results to those observed earlier.

\subsection{Human evaluation: annotation details}
\label{sec:annotation_details}

We provide the details of the annotation required for the human evaluation task (\autoref{sec:generation_annotation}).
We annotate the questions for each combination of subreddit and group category, and we recruit 1 annotator per task via Prolific, with 3 social groups $\times$ 5 subreddits $\times$ 3 annotators = 45 annotators total, and a maximum of 50 questions total for each annotator.
For domain-specific subreddits,
% (\texttt{LegalAdvice}, \texttt{PCMasterRace}, \texttt{PersonalFinance}), 
we recruit annotators based on profession, e.g. annotators who work in the finance industry for \texttt{r/PersonalFinance}.
We pay our annotators \$5 for the task, assuming about 30 minutes per task.
% We provide the ground-truth question, the text-only model question, and one of the social-token model questions in randomized order.
Annotators judged question quality on a 5-point scale based on whether they were answerable, relevant, and understandable.
% from ``Not at all QUALITY'' (1) to ``Very QUALITY'' (5).
The annotators achieved reasonable agreement considering the subjective nature of the task, with Krippendorff's alpha at 0.153 for ``Answerable,'' 0.309 for ``Relevant,'' and 0.23 for ``Understandable'' (compared to 0 for random chance).
% , which seem low but are understandable considering the subjective nature of the task.
% scripts/models/collect_generated_text_eval_data.ipynb

\subsection{Human evaluation: social group prediction}

\label{sec:generation_annotation_social_group}

\begin{table}[t!]
\centering
\small
\begin{tabular}{l r}
Data & Accuracy \\ \toprule
\textbf{Overall} & 47.5 \\ \hline
\multicolumn{2}{c}{\textbf{Social group}} \\ \hline
\texttt{EXPERTISE} & 49.3 \\
\texttt{LOCATION} & 60.8 \\
\texttt{TIME} & 36.9 \\ \hline
\multicolumn{2}{c}{\textbf{Subreddit}} \\ \hline
\texttt{Advice} & 45.6 \\
\texttt{AmItheAsshole} & 48.9 \\
\texttt{LegalAdvice} & 53.3 \\
\texttt{PCMasterRace} & 42.9 \\
\texttt{PersonalFinance} & 47.8 \\ \bottomrule
\end{tabular}
\caption{Human annotation accuracy for group guessing task.}
\label{tab:annotation_question_group_accuracy}
% models/collect_generated_text_eval_data.ipynb#Final-generated-text-eval
\end{table}

We report here the results of the additional annotation task mentioned in \autoref{sec:generation_annotation}.
Following the question quality task, for each post we provide the two social-token model questions
in random order for a group prediction task, where annotators must choose the question that corresponds to a given social group in the category: e.g. ``Which question was more likely to be written by an \textbf{expert} reader?''
We show the results for the group-guessing task in \autoref{tab:annotation_question_group_accuracy}.
Annotators generally had trouble guessing the identity of the social groups except for the \texttt{LOCATION} category, which corresponds with the higher quality ratings reported in \autoref{tab:annotation_question_quality_scores}.
We also find slightly higher guessing accuracy for \texttt{LegalAdvice}, which may be due to intuitive understanding among annotators on what constitutes a difference in social groups for the legal domain (e.g. experts using particular terminology).
The low performance in this task may indicate that human-understandable differences between the questions may be less obvious in individual pairs of questions as compared to the aggregate groups of questions (see differences in \autoref{sec:compare_social_groups}).

\end{document}